\documentclass[letterpaper, 10 pt, conference]{ieeeconf}  %
\pdfoutput=1

\IEEEoverridecommandlockouts                              

\usepackage[backend=bibtex8,
            hyperref=false,
            url=false,
            isbn=false,
            doi=false,
            backref=false,
            style=ieee,
            citestyle=numeric-comp,
            sorting=nyt,
            block=none]{biblatex}
            
\newcommand{\etal}{et al.~}

\renewcommand{\bibfont}{\small}
\usepackage{graphics}
\usepackage[pdftex]{graphicx}
\usepackage{wrapfig}
\DeclareGraphicsExtensions{.pdf,.png,.jpg}
\usepackage{epsfig}
\usepackage[font={footnotesize}]{caption}
\usepackage{subcaption}
\usepackage[rightcaption]{sidecap}
\usepackage{pbox}

\usepackage{bigstrut}
\usepackage{mathtools}
\usepackage{amsmath, amssymb, amscd}
\usepackage{ wasysym } 
\usepackage{amsfonts}
\usepackage{mathptmx} 
\usepackage{gensymb} 
\usepackage{nicefrac}       
\numberwithin{equation}{section} 

\DeclareMathAlphabet{\mathcal}{OMS}{lmsy}{m}{n}
\DeclareSymbolFont{largesymbols}{OMX}{cmex}{m}{n}
\usepackage{textcomp} 

\usepackage{algorithm} 
\usepackage{algorithmicx, algpseudocode} 

\usepackage{array} 
\usepackage{tabularx}
\usepackage{multirow}
\usepackage{multicol}
\usepackage{booktabs}
\usepackage{tabulary}

\usepackage[T1]{fontenc} 
    \usepackage{ae} %
    \usepackage{aecompl}
\usepackage[utf8]{inputenc}
\usepackage[protrusion=true,expansion=true]{microtype}
\usepackage[english]{babel} 
\usepackage{units}
\usepackage{siunitx}
\usepackage{bm}
\usepackage{times} 
\usepackage{xspace}
\usepackage{flushend}
\usepackage{balance} 
\usepackage{csquotes}
\usepackage{makeidx}
\usepackage{blindtext}

\let\labelindent\relax
\usepackage{enumitem}

\usepackage{tikz}
\usetikzlibrary{backgrounds}

\usepackage{ragged2e}
\usepackage{soul} 
\usepackage{subfiles} 

\usepackage[yyyymmdd]{datetime}

\date{\protect\formatdate{1}{1}{2001}}

\usepackage{url}
\makeatletter
\g@addto@macro{\UrlBreaks}{\UrlOrds}
\makeatother
\usepackage{color}
\usepackage{pgfplots}
\pgfplotsset{compat=newest}

\usepackage{hyperref}
\hypersetup{
    colorlinks=true,
    linkcolor=black,
    citecolor=black,
    filecolor=cyan,
    urlcolor=black
}

\usepackage{marginnote}
\usepackage{soul} 

\usepackage[colorinlistoftodos,prependcaption,textsize=small]{todonotes}
\usepackage{regexpatch}
\makeatletter
\xpatchcmd{\@todo}{\setkeys{todonotes}{#1}}{\setkeys{todonotes}{inline,#1}}{}{}
\makeatother

\newcommand{\tocite}[1]{%
\textcolor{red}{[cite:\ifthenelse{\equal{#1}{}}{}{#1}?]}
}

\newcommand{\ignore}[1]{}



\newcommand{\params}{\mathbf{\theta}}

\renewcommand{\theequation}{\arabic{equation}}

\newcommand{\state}{\mathbf{x}}
\newcommand{\control}{\mathbf{u}}
\newcommand{\obs}{\mathbf{z}}
\newcommand{\raw}{\mathbf{D}}

\newcommand{\motionm}{p(\state_{t} \mid \state_{t-1}, \control_{t-1})}
\newcommand{\obsm}{p(\obs_{t} \mid \state_{t})}

\newcommand{\noisem}{\mathbf{q}}
\newcommand{\noiseo}{\mathbf{r}}

\newcommand{\A}{\mathbf{A}}
\newcommand{\Q}{\mathbf{Q}}
\newcommand{\R}{\mathbf{R}}
\title{\LARGE \bf
Multimodal Sensor Fusion with Differentiable Filters}

\author{Michelle A. Lee$^{*}$, Brent Yi$^{*}$, Roberto Mart\'in-Mart\'in, Silvio Savarese, Jeannette Bohg
\thanks{*Equal contribution. All authors are with Stanford Artificial Intelligence Lab (SAIL), Stanford University. {\footnotesize [mishlee, brentyi, robertom, ssilvio, bohg]@stanford.edu}}
\thanks{
This work has been partially supported by JD.com American Technologies Corporation (``JD'') under the SAIL-JD AI Research Initiative. This article solely reflects the opinions and conclusions of its authors and not JD or any entity associated with JD.com.
}%
}

\begin{document}
\maketitle
\thispagestyle{empty}
\pagestyle{empty}

\begin{abstract}

Leveraging multimodal information with recursive Bayesian filters improves performance and robustness of state estimation, as recursive filters can combine different modalities according to their uncertainties.
Prior work has studied how to optimally fuse different sensor modalities with analytical state estimation algorithms.
However, deriving the dynamics and measurement models along with their noise profile can be difficult or lead to intractable models.
Differentiable filters provide a way to learn these models end-to-end while retaining the algorithmic structure of recursive filters.
This can be especially helpful when working with sensor modalities that are high dimensional and have very different characteristics.
In contact-rich manipulation, we want to combine visual sensing (which gives us global information) with tactile sensing (which gives us local information).
In this paper, we study new differentiable filtering architectures to fuse heterogeneous sensor information.
As case studies, we evaluate three tasks: two in planar pushing (simulated and real) and one in manipulating a kinematically constrained door (simulated).
In extensive evaluations, we find that differentiable filters that leverage crossmodal sensor information reach comparable accuracies to unstructured LSTM models, while presenting interpretability benefits that may be important for safety-critical systems.
We also release an open-source library for creating and training differentiable Bayesian filters in PyTorch, which can be found on our project website: \url{https://sites.google.com/view/multimodalfilter}.

\end{abstract}

\section{Introduction}

\label{sec:intro}

\begin{figure}[t]
\centering
\includegraphics[trim={5.0cm 3.0cm 4.8cm 1.0cm},clip,width=0.44\textwidth]{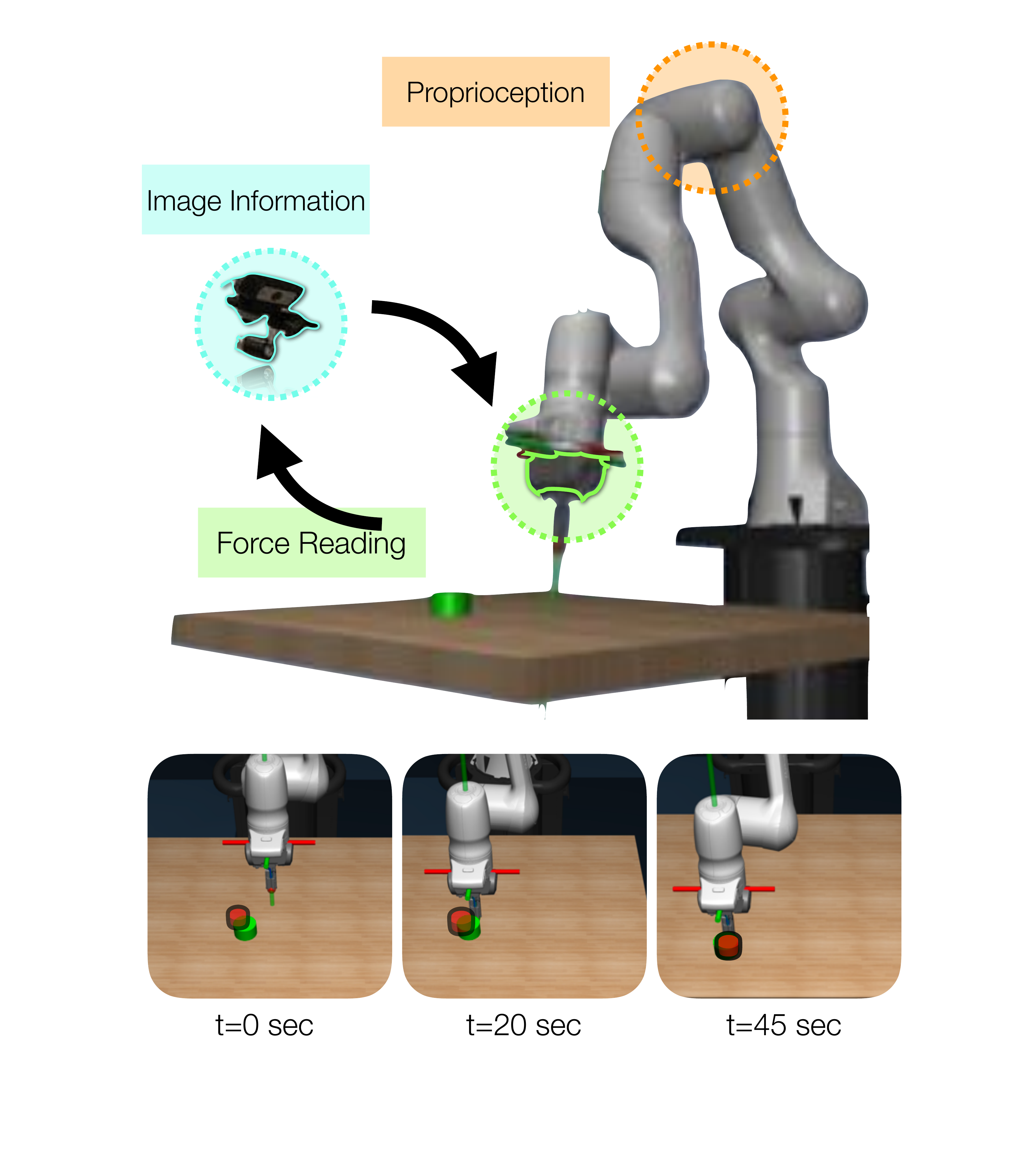}
\caption{
We study the integration of modalities for recursive state estimation with differentiable filters; (top) Information from force sensing (local information), joint encoders and visual sensing (global information) is fused with different architectures, including new crossmodal strategies that use information from one modality to assess the uncertainty of another; (bottom) By combining local and global information during manipulation, the pose of an object (green peg) can be correctly estimated (red overlay).
}
\label{fig1}
\end{figure}

Object manipulation in robotics is inherently multimodal: a robot can observe the object interaction with its cameras, register the motion of its own motors, and sense the forces and torques it applies with its end effector.
The information from these modalities can be used to continuously estimate the state of those elements in the environment that are relevant to the manipulation task.
Since the information from vision, proprioception, and haptics is complementary, using multimodal sensory inputs can help overcome limitations of single modalities due to occlusions, loss of contact, sensor failure, or differing sensor rates.
Fusing these modalities in an optimal manner promises more accurate, robust and reliable manipulation.
This idea has fueled research in sensor fusion and multimodal state estimation for decades~\cite{MultimodalFusion:88,MultimodalFusion:2015,MultimodalFusion:2017}.

Multimodal state estimation for manipulation is challenging.
Visual, proprioceptive and haptic data live in very different sensor spaces that represent different physical properties, have disparate dimensionality, and may arrive at different rates.
Moreover, the models that map these signals to the task-relevant state (e.g. object motion) are often complex and require full knowledge of the environment.
To alleviate the complexity of deriving these models, recently proposed methods~\cite{jonschkowski2018differentiable, haarnoja2016backprop, kloss2018learning} learn the measurement, forward, and noise models from labeled data.
To apply deep neural networks to this learning task, these methods turn the state estimation process into a differentiable procedure that allows end-to-end backpropagation of errors.
While these methods demonstrate improved performance and are more interpretable when compared to completely unstructured LSTMs, this was only shown with a single sensor modality or with sensors with very similar characteristics, such as RGB images and depth. 

To fuse sensor information with different characteristics, an estimator needs to understand how to balance confidence between modalities. 
%
%
We present three types of fusion mechanisms for differentiable filters. The first one fuses data based on uncertainty estimates per modality, i.e. unimodal information. The second ones concatenates unimodal features into a multimodal representations. And a third one uses crossmodal information to weight unimodal state estimates from multiple sensors. We integrate the proposed fusion mechanisms with parametric and non-parametric recursive state estimators.
We compare these multimodal differentiable filters with an unstructured LSTM and evaluate their performance for clean simulated data, noisy simulated data, and real data.

Our study shows that differentiable filters, when compared to more expressive LSTM models, are able to provide important interpretability benefits without sacrificing performance.
In our comparison of fusion mechanisms, we demonstrate that filters that only leverage unimodal information can be brittle and perform inconsistently across tasks.
For filters that leverage crossmodal weights, we demonstrate additional opportunities for interpretability.
By analyzing the outputs of individual components within the proposed crossmodal architectures, we show that these filters allow us to inspect the contributions of individual modalities, while also assessing when information from each modality can be trusted.

\section{Related Work}
\label{sec:tw}

\subsection{Multimodal Fusion}
Traditionally, sensor fusion is performed within a recursive filter by individually modelling each observation and then integrating the resulting information into a common state estimate~\cite{allen1988integrating,doi:10.1177/0278364913497816,hebert2012combined,marton2011combined}.
These methods support the intuition that an optimal combination of information from different modalities improves the overall estimation accuracy. The general schema of these approaches includes a Bayesian filter (see Sec.~\ref{sec:s_mm}) and a multimodal measurement model that quantifies per modality how likely the current measurement is given the predicted state. Other approaches use measurements in a crossmodal manner where information from one sensor modality helps to interpret another~\cite{martin2017cross, martin2017building,cifuentes2016probabilistic}. Each of these methods require the user to identify and define analytical forward and measurement models that may be hard to specify for some dynamical systems or intractable to compute online. In this paper, we reduce the need for predefined models for multimodal and crossmodal fusion by learning both the measurement and the forward model from annotated data with neural networks.

When using probabilistic recursive state estimation with observations from multiple sensors, the information of the modalities can only be optimally integrated if the uncertainty of each sensor has been correctly characterized~\cite{bar1995multitarget,thrun2000probabilistic}. Traditionally, these uncertainties have to be hand-tuned so that the posterior state estimate is close to ground truth. In this paper, we learn the uncertainty of each sensor modality, thereby alleviating handtuning.

Instead of fusing sensors recursively within the filters, Caron \etal proposed to weight state estimates from uni- and multimodal Kalman filters~\cite{caron2006gps} and to switch between measurement models of uni- and multimodal particle filters~\cite{caron2007particle}. However, the weighting mechanisms depend on user-defined and hand-tuned thresholds for each sensor. In our work, we propose and evaluate fusion models that learn to merge state estimates from single modality filters by leveraging crossmodal information. 

\cite{yu2018realtime, lambert2019joint} have proposed a multimodal state estimation framework for planar pushing (similar to two of the three case studies we use in this paper) with factor graphs (iSAM~\cite{iSAM}), however they used fiducials for tracking instead of raw RGB data. While the authors show that iSAM provides more accurate and robust state estimates, it still requires careful specification of modality-specific cost functions as well as forward and measurement models, which we learn from data. 

\subsection{Differentiable filters}
Differentiable filters provide an approach for learning forward and measurement models from data while retaining the algorithmic structure of a recursive Bayes filter.
Differentiable filters can thus be advantageous for systems whose dynamics and sensor observations are hard to model analytically, while making it possible to retain the interpretability of state representation and uncertainty that is often vital for safety-critical systems.
Jonschkowski~\etal\cite{jonschkowski2018differentiable} and Karkus~\etal\cite{karkus2018particle} each proposed differentiable versions of the particle filter~\cite{thrun2000probabilistic}, as applied to simple localization tasks. Their methods learn an estimator that shows improved results over an unconstrained LSTM. Similarly, Haarnoja~\etal\cite{haarnoja2016backprop} proposed a differentiable version of the EKF. The authors reach similar results after testing on toy visual tracking and real-world localization tasks.
In addition, other lines of research have proposed to use recurrent neural networks to learn latent state representations~\cite{ma2019particle, hafner2019learning, le2017auto, igl2018deep}.
All methods above explored how to combine filtering techniques with learning approaches, but they do not systematically study how to fuse information from different sensor modalities. 
In this paper, we investigate new differentiable filtering architectures to fuse information from the heterogenous sensor modalities of vision, touch, and proprioception.

\section{Background on Filtering}
\label{sec:s_mm}
We consider the problem of estimating the state $\state$ of a system from a sequence of (multimodal or unimodal) observations $\obs$ and control inputs $\control$. We represent our knowledge and uncertainty about our estimate with a distribution over the current state $\state_t$ conditioned on all the previous observations $\obs_{1:t}$ and control inputs $\control_{1:t}$. We denote this distributions as belief $bel(
\state_t) = p(\state_t \mid \control_{1:t}, \obs_{1:t})$. One solution to compute this belief are Bayes filters.

Traditionally, developing a Bayes filter requires analytically formulating a forward model to predict the next state based on current state and control input, $\motionm$, a measurement model, $\obsm$, to compute the likelihood of an observation given a state, and the noise associated with predictions and observations. Formulating these models and quantifying the noise often requires making strong assumptions on the properties of the underlying system.
A way to avoid making these assumptions is to extract this information directly from labeled data of observations and ground truth states. This can be achieved with the recent family of differentiable Bayes filters. Here, we study differentiable filters for the special case of fusing heterogeneous sensory data from multiple modalities. In this section, we will first give a brief summary of Bayes filters before we provide a unified notation for differentiable filters. For a more in-depth discussion of Bayes filters, we refer to~\cite{thrun2000probabilistic}. 

\subsection{Bayes Filters}
The Bayes filter algorithm provides an optimal solution for state estimation in a system that follows the Markov assumption and in which observations are conditionally independent. In this filter, the belief is updated in two steps. In the prediction step, a motion model $\motionm$ is used to predict the belief $\overline{bel}(\state_t)$ about the current state $\state_t$ given the previous state $\state_{t-1}$ and control input $\control_{t-1}$
\begin{equation}
\label{eq:predict}
    \overline{bel}(\state_t) = \int \motionm bel(\state_{t-1}) d \state_{t-1}
\end{equation}
In the update step, we correct this initial prediction given a sensory observation $\obs_t$ plugged into a measurement model $\obsm$ that describes the likelihood of this observation given the predicted state:
\begin{equation}
\label{eq:update}
    bel(\state_t) = \eta \obsm \overline{bel}(\state_t)
\end{equation}
where $\eta$ is a normalization factor.

The Kalman filter~\cite{kalman1960new} is the optimal estimator for a system with linear models and Gaussian noise. In robotics, however, most systems of interest have non-linear models and may follow a more complex non-Gaussian distribution. For these cases, there are non-linear Bayes filters that make different approximations to estimate the belief $bel(\state_t)$. 

Of these, the most widely used include the Extended Kalman Filter (EKF) and the Particle Filter~\cite{thrun2000probabilistic}. The EKF is used when the involved system models are not linear but linearizable through a Taylor expansion. Researchers resort to particle filters when the non-linearities of the system cannot be linearized or when the underlying state distribution cannot assumed to be Gaussian, e.g. when distributions have multiple peaks. In this work we will study how to fuse information from multiple sensor modalities using differentiable versions of the EKF and particle filter.

\subsection{Differentiable Filters}
\label{ss_df}
For complex physical systems, it is often a challenge to formulate a dynamics and observation model that is both accurate and tractable to compute. Recently, differentiable versions of the most popular non-linear filtering algorithms have been proposed such that dynamics and observation models along with their noise parameters can be learned. 


\subsubsection{Extended Kalman Filter}
The EKF allows non-linear forward and observation models but still assumes that the state follows a Gaussian distribution such that $bel(\state_t) \sim N(\mu_t, \Sigma_t)$. Specifically, we assume that the system dynamics and measurements follow the following nonlinear functions: 
\begin{align}
\state_t &= f(\state_{t-1}, \control_{t-1}, \noisem_{t}) \\
 \obs_t &= h(\state_t, \noiseo_t)
\end{align}
where the random variables $\noisem$ and $\noiseo$ are the process and observation noise. In the EKF, the prediction step is as follows
\begin{align}
\label{eq:mu} 
\hat{\mu}_t &=  f(\mu_{t-1}, \control_{t-1}, 0)\\
\label{eq:sigma}
\hat{\Sigma}_t &= \mathbf{A}_{t-1}\Sigma_{t-1} \mathbf{A}_{t-1}^T + \mathbf{Q}_{t-1} 
\end{align}
where $\A_t$ is the Jacobian $\frac{\partial f(\mu_t, \control_t, 0)}{\partial \mu_t}$ and $\Q$ the covariance of the process noise $\noisem \sim N(0,\Q)$ which we assume to be Gaussian with zero mean. Therefore, $\overline{bel}(\state_t) \sim N(\hat{\mu}_t, \hat{\Sigma}_t)$.

The update step is as follows:

\begin{align}
\label{eq:kalmangain}
\mathbf{K}_t &= \hat{\Sigma}_t\mathbf{H}_t^T (\mathbf{H}_t \hat{\Sigma}_t \mathbf{H}_t^T + \mathbf{R}_t)^{-1} \\
\label{eq:mu_update}
\mu_t &= \hat{\mu}_t + \mathbf{K}_t (\mathbf{z}_t - \mathbf{H}_t \hat{\mu}_t)  \\
    \label{eq:sigma_update}
\Sigma_t &= (\mathbf{I}_n - \mathbf{K}_t \mathbf{H}_t ) \hat{\Sigma}_t 
\end{align}
where $\mathbf{H}_t$ is the Jacobian $\frac{\partial h(\mu_t, 0)}{\partial \mu_t}$ and $\R$ the covariance of the measurement noise $\noiseo \sim N(0,\R)$ which we assume to be Gaussian with zero mean.

For a differentiable version of the EKF, we implement the dynamics model in Eq.~\ref{eq:mu} with a {\em Multi-Layer Perceptron} (MLP) with trainable weights $\params$: $f_\params(\mu_{t-1}, \control_{t-1}, 0)$.
Learning a true measurement model that maps states to expected observations in our raw sensor space is an underdetermined problem and prone to overfitting. We follow~\cite{haarnoja2016backprop} and instead learn a discriminative virtual sensor.
Our virtual sensor $g_\params(\raw)$ uses as input raw sensory data $\raw$ and outputs a vector $\mathbf{z}_t$ containing observations of the full state $\state$ or parts of the state.
Therefore, the measurement model and its Jacobian $\mathbf{H}$ are either identity or selection matrices.
Similar to the dynamics model, we implement the virtual sensor $g_\params(\raw)$ with an MLP with trainable weights $\params$.

\begin{figure}[!t]
\centering
  \includegraphics[width=0.8\linewidth]{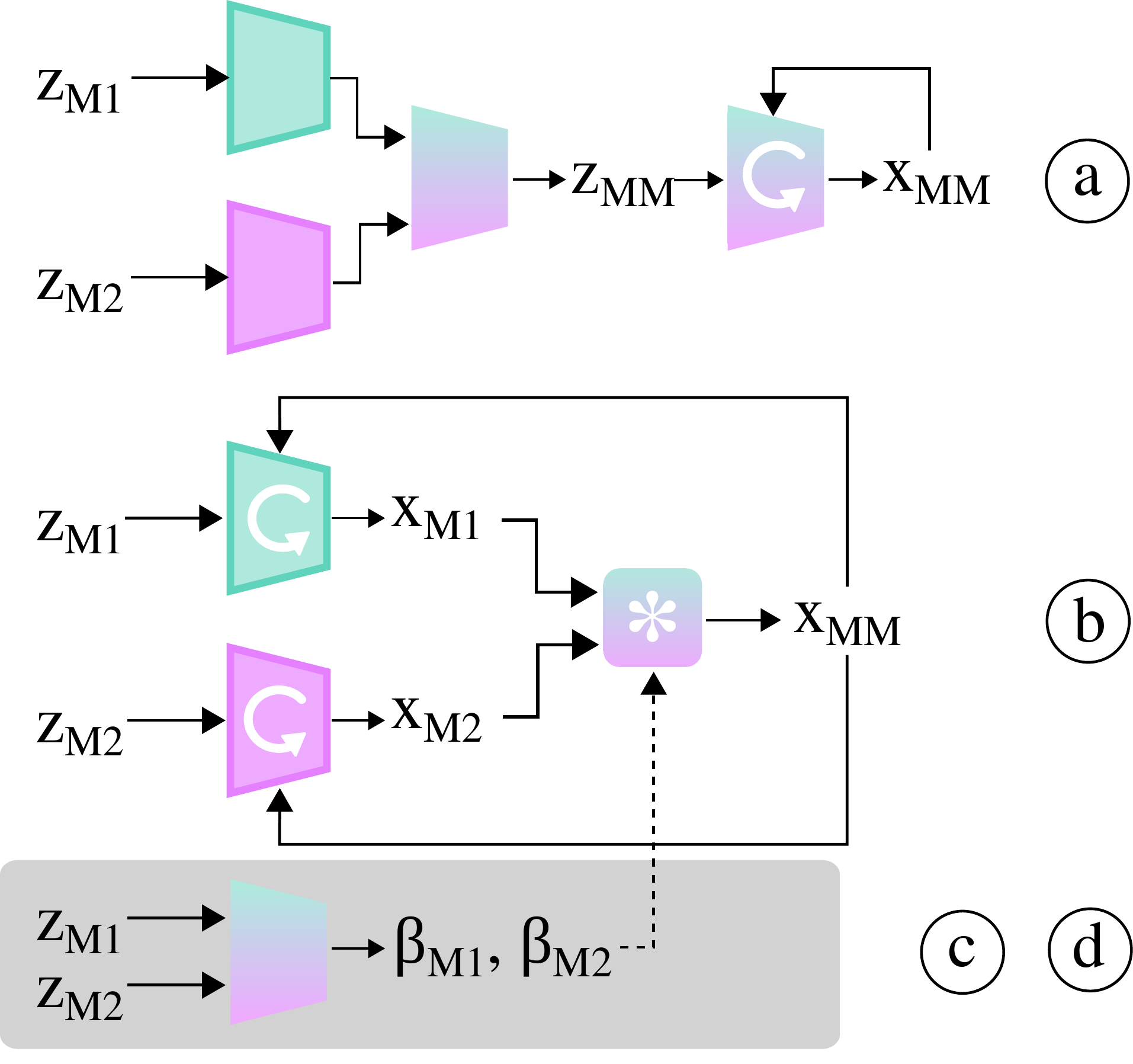}
    \caption{Multimodal Architectures: \textbf{a) Fusion of Features}: modalities are processed and merged into a multimodal feature for state estimation; \textbf{b) Unimodal Weighted Fusion}: estimates from two unimodal recursive filters are merged based on their uncertainty; c) and d) \textbf{Crossmodal Weighted Fusion}: both modalities are used to learn coefficients to fuse the unimodal estimates; the coefficients are used for \textbf{c) weighted average of states} or \textbf{d) weighted measurement model}}
  \label{fig:fusingstrategies}
\end{figure}
\subsubsection{Particle Filter}
While the EKF assumes a Gaussian distribution for the underlying distribution of the system state, a particle filter can model arbitrary distributions by representing them with a set of particles.
Concretely, we will learn a measurement model in the form of a neural network that outputs the log-likelihood of the observation predicted for each particle.
Specifically, we approximate the belief with a set of particles $\mathcal{X}_t = \state_t^{[1]},\state_t^{[2]}, \cdots , \state_t^{[n]}$ with weights $w_t^{[1]},w_t^{[2]}, \cdots , w_t^{[n]}$.
Like any Bayes filter, this non-parametric filter has a predict and update step. The prediction step in Eq.~\ref{eq:predict} is implemented by sampling a random perturbation for each particle from a generative motion model: 
\begin{align}
\forall_i : \state_t^{[i]} \sim p(\state_t \mid \control_{t-1}, \state_{t-1}^{[i]})    
\end{align}
The update step in Eq.~\ref{eq:update} is implemented by setting the weight $w_t^{[i]}$ of each particle equal to the likelihood of the current measurement $\obs_t$ being generated by the predicted state $\state_t^{[i]}$ of this particle:
\begin{align}
\forall_i : w_t^{[i]} \sim p(\obs_t \mid \state_t^{[i]})    
\end{align}
The particle set is then re-sampled proportionally to the weight of each particle. 
Note that a virtual sensor is needed only for the EKF and not the particle filters. For the particle filters, $\obs_t$ equals the raw sensory data $\raw$ without any processing.


\subsubsection{LSTM}
A recurrent filter resembles the structure of a long-short term memory: an internal state that is recurrently merged with the latest input signal to generate an updated output.
Therefore, we consider LSTM architectures as a baseline to differentiable filters, which we call \texttt{LSTM Baseline}.
In contrast to a differentiable filter, there is no explicit separation between prediction and update steps in an LSTM. There are also no explicit measurement or forward models. An LSTM also does not make the Markov Assumption. Previous studies~\cite{karkus2018particle,haarnoja2016backprop,jonschkowski2018differentiable} demonstrated that differentiable architectures that leverage the algorithmic structure of a Bayes filter lead to faster learning and better generalization than generic LSTM architectures.




\section{Architectures for Multimodal Fusion}
\label{ss_fus}

We study different strategies to integrate the information from multiple sensor modalities into a coherent state estimate. We now describe these strategies as visualized in Fig.~\ref{fig:fusingstrategies}.

\subsection{Feature Fusion}

One strategy to fuse information from multiple modalities in a differentiable filter is to first extract features from each modality separately, and then estimate recursively the state using the fused unimodal features (Fig.~\ref{fig:fusingstrategies}, a). This is achieved by a separate encoder network for each modality, and a fully connected network to integrate the unimodal features into a multimodal one. The multimodal feature is then used as an observation in the recurrent filter architecture. We call our filters \texttt{Feature Fusion EKF} and \texttt{Feature Fusion PF}.

\subsection{Unimodal Weighted Fusion}
\label{ss_uniwa}
This architecture (Fig.~\ref{fig:fusingstrategies}, b) is a fusing procedure for multiple state estimates that are each normally distributed. In this case, we assume we have two unimodal EKF filters that provide independent state estimates $bel(
\state_t^{M_1}) \sim \mathcal{N}(\mu^{M_1}_t, \Sigma^{M_1}_t)$ and $bel(
\state_t^{M_2}) \sim \mathcal{N}(\mu^{M_2}_t, \Sigma^{M_2}_t)$. We fuse the two unimodal beliefs by multiplying the two distributions to produce a normally distributed multimodal belief $bel(\state_t^{MM})$:
\begin{align}
    bel(\state_t^{MM}) &= \mathcal{N}(\mu^{MM}_t, \Sigma^{MM}_t)\\
    \mu^{MM}_t &= \frac{(\Sigma^{M_1}_t)^{-1}\mu^{M_1}_t+(\Sigma^{M_2}_t)^{-1}\mu^{M_2}_t}{(\Sigma^{M_1}_t)^{-1}+(\Sigma^{M_2}_t)^{-1}}\\
    \Sigma^{MM}_t &= ((\Sigma^{M_1}_t)^{-1}+(\Sigma^{M_2}_t)^{-1})^{-1}
\end{align}
This product of two Gaussians is equivalent to the Product of Experts~\cite{hinton1999products}.
In this architecture, there is no crossmodal information flow since the modalities are assumed independent of each other: the information from one does not help estimating or assessing the uncertainty of the other. We refer to the resulting model as \texttt{Unimodal Fusion EKF}. 

\subsection{Crossmodal Weighted Fusion}
\label{ss_wa}
In this architecture, we also assume that there are unimodal filters providing individual state estimates. Similar to the \texttt{Unimodal Fusion EKF}, we assume that the estimates are normally distributed and independent. The integration of their estimates is goverened by the coefficients $\beta_{t}^{M_1}$ and $\beta_{t}^{M_2}$, which weight the contributions of each unimodal value (estimated from observations of sensors $M1$ or $M2$) into a fused estimate. 
Each coefficient is inferred from the information contained in the multimodal signal. In this way, information from one modality is used to assess the uncertainty about the other and vice versa, creating a crossmodal information flow. This architecture assumes that the final and each unimodal belief can be faithfully represented by a Gaussian. We apply this fusion architecture to integrate information from several differentiable Extended Kalman filters and we call the resulting model \texttt{{Crossmodal Fusion EKF}}.

In our implementation, $\Vec{{\beta}}^{M_i}_t$ consists of elements ${\beta}^{M_i}_{t, 1},{\beta}^{M_i}_{t, 2}, \hdots {\beta}^{M_i}_{t, n}$, where $n$ is the dimensionality of the state space. The weighting of the covariance matrix is done through a positive semi-definite matrix $\mathbf{B}^{M_i}_t \in \mathbb{R}^{n \times n}$, defined as:

\[\mathbf{B}^{M_i}_t =
\begin{bmatrix}
\beta^{M_i}_{t,1}& \hdots & \beta^{M_i}_{t,n}
\\ \vdots & & \vdots
\\ \beta^{M_i}_{t,1}& \hdots & {\beta}^{M_i}_{t,n}
\end{bmatrix} ^T \begin{bmatrix}
\beta^{M_i}_{t,1}& \hdots & \beta^{M_i}_{t,n}
\\ \vdots & & \vdots
\\ \beta^{M_i}_{t,1}& \hdots & {\beta}^{M_i}_{t,n}
\end{bmatrix}. 
\]
The integration is then a weighted average as follows:
\begin{align}
    \mu^{MM}_{t}&=  \frac{ \Vec{\beta}^{M_1}_{t}\odot \mu^{M_1}_{t} + \Vec{\beta}^{M_2}_{t}\odot \mu^{M_2}_{t}}{\Vec{\beta}^{M_1}+\Vec{\beta}^{M_2}_{t}} \\
    \Sigma^{MM}_t &= \frac{ \mathbf{B}^{M_1}_{t} \odot \Sigma^{M_1}_{t} 
    + \mathbf{B}^{M_2}_{t} \odot \Sigma^{M_2}_{t} }
    {\mathbf{B}_{t}^{M_1}+\mathbf{B}^{M_2}_{t}}
\end{align}
Note that in this case, the superscript \textit{$M_1$} and \textit{$M_2$} does not indicate the originating modality but the estimated state that it is applied to. 

\subsection{Unimodal Weighted Measurement Models}
\label{ss_uniwm}

In this architecture, we assume unimodal measurement models $\obsm^{M_1}$ and $\obsm^{M_2}$ that provide unimodal weights $w_t^{M_1,[i]}$ and $w_t^{M_2,[i]}$ for each particle.
Unlike in Section \ref{ss_uniwa}, where we combine estimated states from each filter, we instead combine the likelihoods that are the outputs of each measurement model. One way to do this is with a mixture model, where we compute our particle weights as the sum of our unimodal weights:

\begin{align}
    w_t^{[i]} = w_t^{M_1,[i]} + w_t^{M_2,[i]}
\end{align}

Using the standard particle filter dynamics update and resampling procedures, we can apply this architecture to integrate estimates from a particle filter with a single dynamics model and multiple unimodal measurement models.
We call the resulting model \texttt{Unimodal Fusion PF}.

\subsection{Measurement Models with Learned Crossmodal Weights}
\label{ss_wm}

Finally, we can also add a crossmodal information flow to the fusion of our unimodal measurement models assumed in Section \ref{ss_uniwm}. 
We do this by learning a mixture weighting model that generates non-negative scalar coefficients $\beta_{t}^{M_1}$ and $\beta_{t}^{M_2}$ from the full multimodal observation input.
This is a differentiable generalization of the discrete measurement model switching method employed by~\cite{caron2007particle} for fusing global and local sensor information in a land vehicle positioning problem. The weight of each particle is therefore set to:
\begin{align}   
    w_t^{[i]} = \beta_{t}^{M_1} * w_t^{M_1,[i]} + \beta_{t}^{M_2} * w_t^{M_2,[i]}
\end{align}

We call the resulting model \texttt{Crossmodal Fusion PF}.

\section{Implementation}
\label{sec:impdetails}

\subsection{Network Architectures}
\label{sec:network}


For the differentiable EKF and differentiable PF estimators, we use a shared \textbf{dynamics model} with terms parameterized by trainable weights $\theta$:
\begin{align}
    \state_{t} &= \state_{t - 1} + f_{1,\theta}(\state_{t - 1}, \control_{t - 1}) \cdot \sigma(f_{2,\theta}(\state_{t - 1}, \control_{t - 1}))
    \label{eq1}
\end{align}
where $f_{1,\theta}$ is a state update vector of the same dimension as $\state_t$, $f_{2,\theta} \in \mathbb{R}$ is a scalar gating/scaling term, and $\sigma$ is the sigmoid function $\sigma(z) = (1 + e^{-z})^{-1}$. 
Separating the relative state update into these two terms allows the network to independently learn the direction and magnitude of the state update.
For the particle filter, we maintain the diversity of particles by injecting additive Gaussian noise in our state space after each dynamics udpate.

The network that outputs $f_{1,\theta}$ and $f_{2,\theta}$ feeds the inputs $\state_{t - 1}$ and $\control_{t - 1}$ through three-layer encoders.
The outputs are concatenated and passed to a seven-layer set of shared layers.

The \textbf{measurement models} of the differentiable PFs are trained using as inputs observations (images, F/T signals and/or end-effector positions) and states (object position on the table) as stored in each particle. The output of the measurement model is the log-likelihood of these observations given the state of each particle.
Image inputs are encoded with a set of 2D convolutions, while proprioception and haptics are processed with standard fully-connected layers. The encoded features are concatenated, and then fed to a shared series of output layers.

The \textbf{virtual sensor model} architectures are nearly identical to our particle filter measurement models, but instead of outputting a log likelihood, the networks output an estimated state and state covariance for our EKF.



\begin{figure}[ht!]
\centering

\includegraphics[ 
width=.99\linewidth]{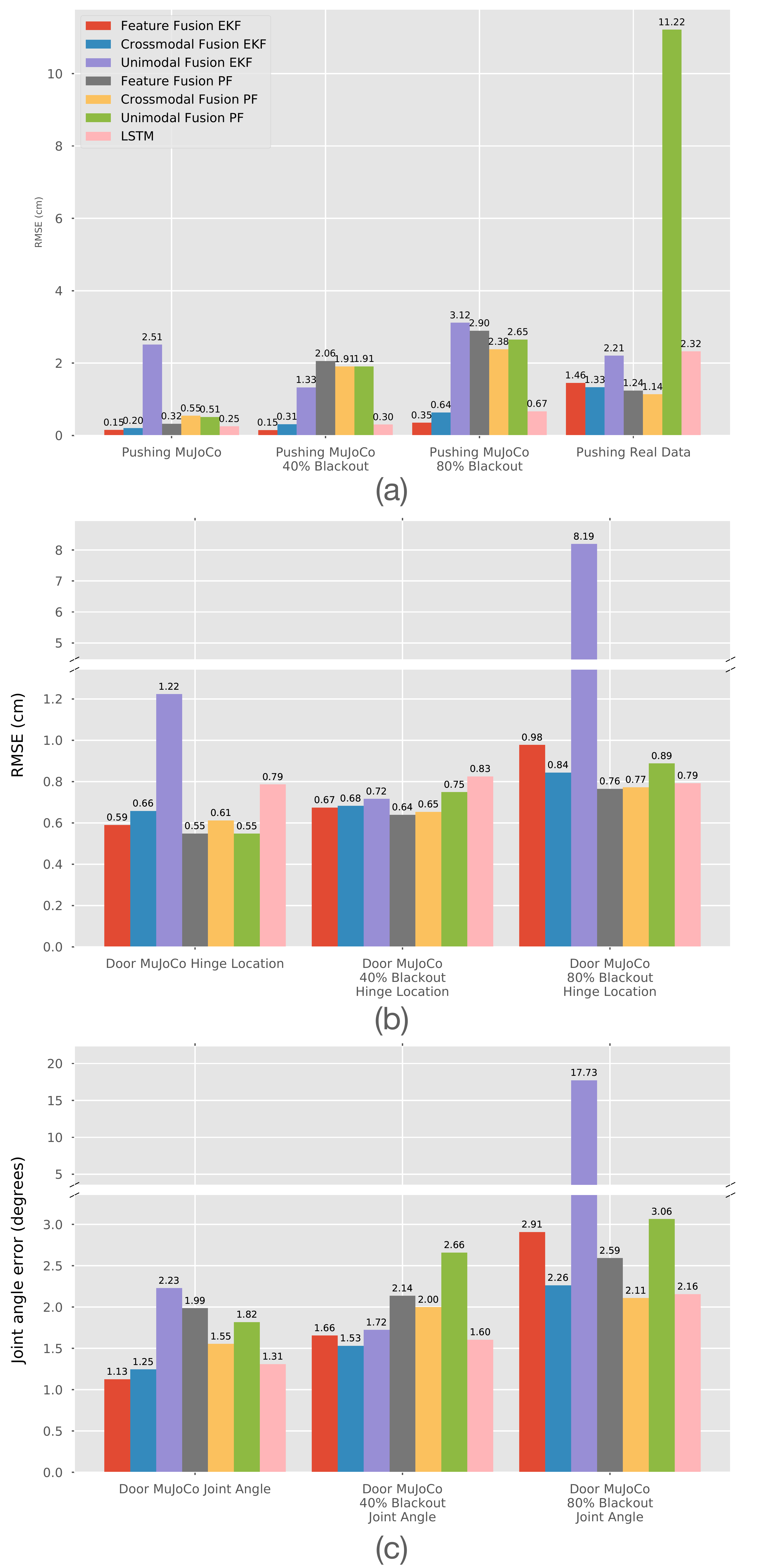}
\caption{We show the results of our tasks: pushing and door manipulation. In (a), RMSE error in centimeters for each filter in our validation set for MuJoCo pushing simulation data, MuJoCo pushing simulation data with 40\% and 80\% image blackout, and real world pushing. In (b) and (c) respectively, RMSE error in centimeters and joint angle error in degrees for each filter in our validation set for the MuJoCo door simulation data and MuJoCo simulation data with 40\% and 80\% image blackout.}
\label{fig:error_all}
\end{figure}

In the \textbf{crossmodal weight models}, the inputs are also observations (images, F/T signals, and/or end-effector positions), and the outputs are the learned crossmodal fusion coefficients used to balance estimates from each modality. Each modality is run through an encoder with the same architecture as our measurement/virtual sensor models, and concatenated before being fed to the output layers that produce our final weights.

The \texttt{LSTM Baseline model} is designed to have a similar architecture and parameter count as the rest of our estimators, with the same encoder architectures for each modality and control inputs.
Encoded features from each modality and action inputs are concatenated and passed through a series of shared layers; in contrast to our crossmodal weight models, these layers are terminated with two LSTM layers.
The output of the final LSTM layer is then mapped through an additional fully-connected layer before an affine mapping is applied to produce the state estimate.

We use ReLU activations and ResNet-style skip connections~\cite{he2015deep} throughout all of our networks.

To ensure that comparisons are fair, we ran an architecture search for each model.
For the image encoders, we tried three architecture variations for mapping outputs of our 2D convolutions to our fully-connected layers: (a) a simple flatten operation, (b) spatial softmaxes, and (c) mean pooling. 
We found the simple architecture (a) to perform well for models that outputted scalar weights and log-likelihoods, while option (c) generalized best for networks that directly regressed XY coordinates.
We additionally explored varying network widths and LSTM hidden state dimensions; final results are reported using 64 units for all fully-connected layers and a hidden state size of 512 for each LSTM layer.

\subsection{Training Procedure}

We first pre-train the dynamics and measurement models before fine-tuning them in an end-to-end manner.
The \textbf{dynamics models} (Sec.~\ref{ss_df}) are pre-trained to minimize single-step prediction errors, and then for $4$, $8$, and finally $16$-step prediction errors.

The \textbf{particle filter measurement models} are pre-trained to predict the observation-conditioned log probability density function of a multivariate Gaussian centered at the ground-truth state for our particle filter.
Likewise, the \textbf{EKF virtual sensor models} are pre-trained to generate observation-conditioned predicted states.
Measurement uncertainties are learned exclusively end-to-end, and not pre-trained.

After pre-training, we use backpropagation through time to train each of our state estimation models end-to-end until convergence, over subsequences of increasing length. $30$ particles are used for training the particle filter. Optimization was done using the Adam optimizer on NVIDIA Quadro P4000, Tesla K80, and Tesla V100 GPUs.

\section{Experimental Evaluation}
\label{sec:experiments}

\begin{figure*}[ht!]
\centering
\includegraphics[ width=.90\linewidth]{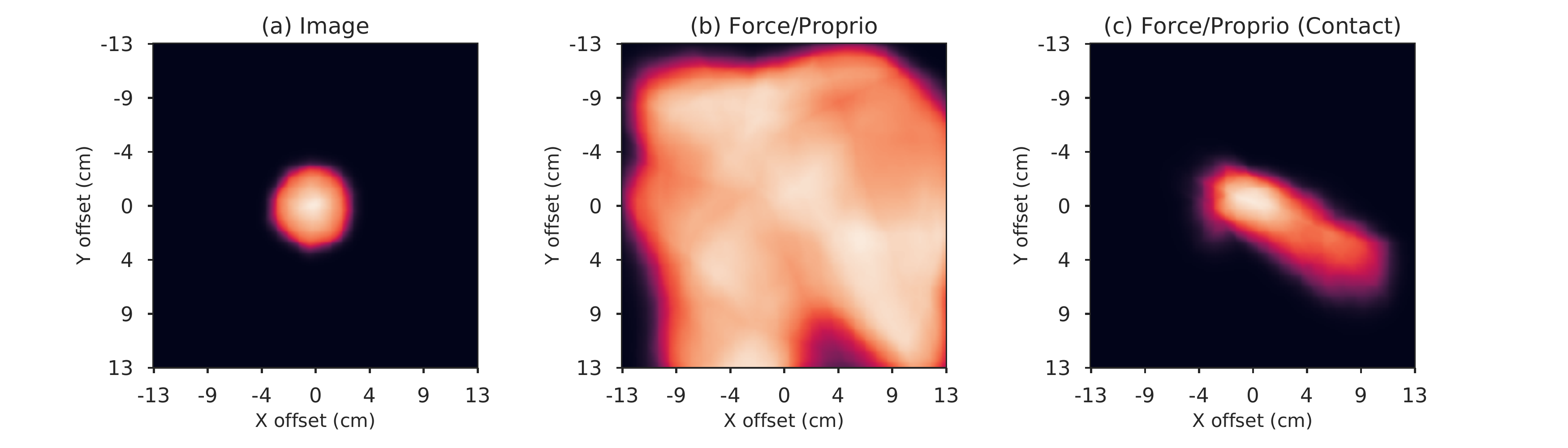}
\caption{
Examining outputs from unimodal measurement models within the proposed fusion architectures can help us understand how our neural networks are fusing each sensing modality.
Using a \texttt{Crossmodal Fusion PF} model trained end-to-end on the \texttt{Mujoco door dataset}, we plot how the likelihood of a state changes as a function of positional offsets from the ground-truth door hinge position: \textbf{(a)} using the unimodal image measurement model, \textbf{(b)} using the force/proprioception model on a timestep before the robot has made contact with the door, and \textbf{(c)} using the force/proprioception model after the robot gripper has made contact with the door.
}
\label{fig:pf_interpret}
\end{figure*}

We study the fusion architectures described in Sec.~\ref{ss_fus} to estimate the state in three experimental scenarios: pushing a planar object in simulation, pushing a planar object in the real world, and opening a door in simulation.
The simulation experiments allow us to evaluate the fusion architectures in a controlled and noise-free setup, while the real data evaluation allows us to test the robustness of the fusion techniques in a realistic domain.
In both simulation and the real world, we first generate a dataset of sensor data and ground truth states, use part of the dataset to train our differentiable models and a different part to test the resulting models.



\subsection{Experimental Setup}
In the planar pushing tasks, we estimate the 2D position of the unknown object on a table surface, $\state_t = (x_t,y_t)$, while the robot intermittently interacts with the object.
In the door manipulation task, we estimate the 2D position of the door's revolute joint and its joint angle, $\state_t = (x_t,y_t,\theta_t)$, while the door is being opened.
In these scenarios, we compare the performance of the aforementioned fusion architectures when using as input vision (images), haptics (force-torque sensor readings), and proprioception (encoder values we use to infer the end-effector pose).
For each of our \textit{Crossmodal Fusion} and \textit{Unimodal Fusion} filters, we consider the image as one weighted modality and force/proprioception as another.

For the simulation experiments, we collect data with MuJoCo~\cite{todorov2012mujoco}.
We create a Mujoco pushing dataset consisting of 1000 trajectories with 250 steps at \SI{10}{\hertz}, of a simulated Franka Panda robot arm pushing a circular puck.
The pushing actions are generated by a heuristic controller that tries to move the end-effector to the center of the object.


For the evaluation with real-world data, we use the three ellipse shapes from the MIT pushing dataset~\cite{yu2016more}, which consists of more than a million real robot push sequences on eleven different objects with four different surface materials. We use the tools described in Kloss~\etal~\cite{kloss2017combining} to obtain additional annotations for the remaining state component, for rendering gray-scale images, and for stitching together 1000 pushing trajectories with 45 steps at \SI{18}{\hertz}. 

Finally, we collect a dataset for the door task in Mujoco that consists of 600 trajectories with 800 steps at \SI{10}{\hertz} each, of the Franka Panda robot pushing and pulling a kinematically constrained door object.
Actions are generated by a heuristic controller that pushes and pulls on the door in random directions.
Half of the trajectories are recorded with the gripper of the robot closed around the handle of the door; the other half is recorded without directly grabbing the handle.

In all experiments, the multimodal inputs are gray-scaled images (1 x 32 x 32) from an RGB camera, forces (and binary contact information) from a force/torque sensor, and the 3D position of the robot end-effector. While wrench signals in the real-world dataset only have 3 dimensions, force in the x and y direction, and torque in the z direction ($F_x$, $F_y$, and $\tau_z$), we use all six dimensions of force and torque in the simulation dataset.  

To see how each filter adjusts to sensor failure and noise, we randomly black out the input images with probabilities of 0.4 and 0.8 in the simulation datasets for pushing and the door task. For evaluating each filter, we split the trajectories into train and test sets. We use 10 trajectories from the simulated pushing task, 50 trajectories from the real-world data pushing task, and 20 trajectories for the door opening task.

\subsection{Experimental Results}

\subsubsection{Pushing Task}
In the pushing task, we evaluate each filter by comparing the root mean squared error (RMSE) of the estimated X and Y location of the pushed object.
We show the results of each filter in Fig.~\ref{fig:error_all}.

Notably, we found that the \texttt{Unimodal Fusion} filters were brittle and inconsistent.
The \texttt{Unimodal Fusion EKF} performs worse on the blackout-free dataset than on the dataset with 40\% blackout images, and \texttt{Unimodal Fusion PF} had an average of 11cm error on the real pushing dataset. 

On the other hand, all models that utilized crossmodal information (the \texttt{Crossmodal Fusion} filters, \texttt{Feature Fusion} filters, and \texttt{LSTM}) were able to learn how to estimate the location of the object with less than 3cm of error in the simulated dataset and less than 2.5cm on the real dataset.

While the EKF and LSTM estimators perform better than the particle filters on our simulation dataset, we notice that this gap is completely closed on the real-world dataset.
We speculate that this is because the particle filter is better at handling discontinuous interactions: our simulation dataset consists of only a single pushing motion per trajectory, while our real-world dataset repeatedly makes and breaks contact.

%

\subsubsection{Door Task}

In the Mujoco door dataset, each filter is trained to estimate a 3-DoF state with different characteristics: the $x$ and $y$ coordinates of the location of the door hinge (stationary within a trajectory) and the hinge angle (changing based on robot's actions).
We evaluate the filter performance on the door task with two metrics: the absolute error of the estimated hinge angle and the RMSE error of the hinge's location (we assume the hinge joint pointing in the $z$ direction).
We show our results in Fig.~\ref{fig:error_all}.

With the exception of \texttt{Unimodal Fusion EKF}, we see comparable results across all models for the door estimation task. For joint angle estimation, \texttt{Crossmodal Fusion} models had slightly better performance than \texttt{Feature Fusion} models. \texttt{LSTM} models remain fairly consistent with both joint angle and hinge location estimations, across all datasets. 


\subsection{Discussion}

\begin{figure}[t]
\centering

\includegraphics[trim={1.0cm 0.5cm 1.0cm 1.0cm},clip,
width=.99\linewidth]{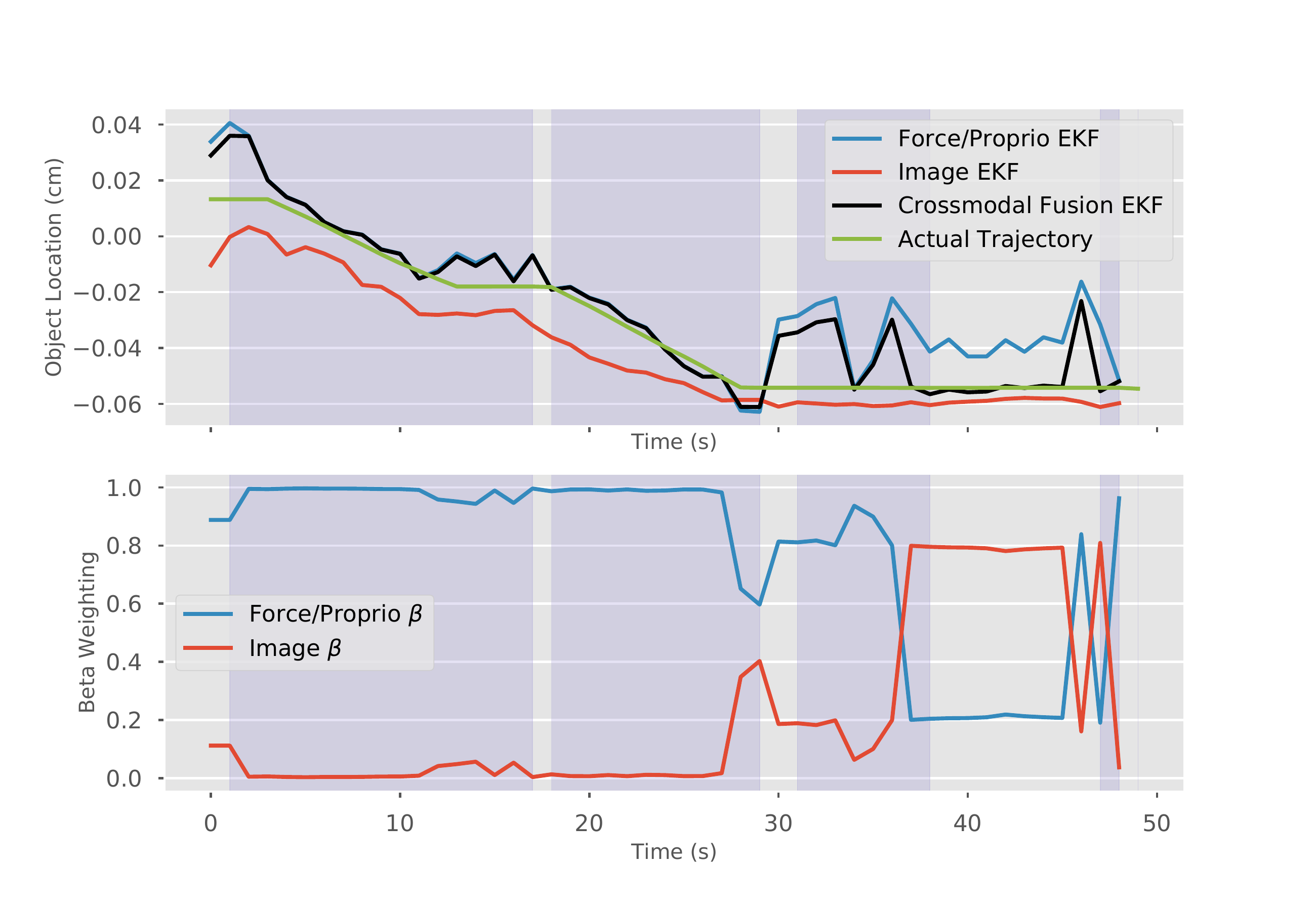}
\caption{For the real pushing task, we show the estimated states of the filters (top) and the weights for the unimodal filters (bottom) from the \texttt{Crossmodal Fusion EKF}. Contact is indicated by the purple background. The \texttt{Crossmodal Fusion EKF} learns to weight the force and proprioceptive unimodal EKF when the robot is touching the object, and switch to the image unimodal EKF when the robot breaks contact with the object. }
\label{fig:ekf_interpret}
\end{figure}

In the \texttt{Crossmodal Fusion} filters, we can inspect the outputs of the weighting models to understand how each modality is balanced. 
Fig.~\ref{fig:ekf_interpret} depicts the predictions from each modality and the learned weights for the fusion of information by the \texttt{Crossmodal Fusion EKF} on the real-world pushing dataset.
We observe that the filter relies more on the information from proprioceptive and force-based modalities when the robot is in contact and pushing the object.
When the robot stops pushing the object or loses contact, the filter learns to put more weight on information from the image-based EKF and less on the proprioception/haptics-based EKF. 

When we use a \texttt{Crossmodal Fusion} filter, we can also examine outputs from each unimodal measurement model to help us understand what the filter is learning from each sensing modality.
Fig.~\ref{fig:pf_interpret} depicts the hinge location likelihood from each unimodal measurement model in an end-to-end trained \texttt{Crossmodal Fusion PF}.
From these visualizations, we observe that while the image input contains information to roughly infer both the $x$ and $y$ coordinates of the location of the hinge (placed at the origin in this case), the force/proprioception signal only reveals that the hinge location should be along the axis between the true hinge location and the end-effector location.
Given the combined information of proprioception and force/torque in one timestep, the hinge location cannot be inferred unequivocally, only its distribution along the connecting line.
By fusing likelihoods from these two unimodal measurement models, the learned filter is able to produce a better estimate of the position than it could if it only had access to one of them.

We expect the interpretability benefits of the proposed \texttt{Crossmodal Fusion} architectures to be especially advantageous for building estimators deployed in settings where safety and reliability are critical.
In addition to using the belief uncertainty intrinsic to a Bayesian filter to inform decision-making, we can also use our understanding of how the estimator is interpreting each input modality to better diagnose why, how, and where in our estimation pipeline failures happen.
By selectively freezing weights in parts of the estimator --- for example, the image filter if we discover an issue with our force/proprioception filter --- we can then patch these failures in an end-to-end manner while minimizing unintended impacts.

Similar to the findings of Kloss~\etal\cite{kloss2018learning}, we find that the performance of each type of multimodal filter depends heavily on the task, with no unique best performing filter.

\section{Conclusion} 
\label{sec:conclusion}
We presented a study of fusion architectures within the field of differentiable recursive filters for state estimation.
For our experiments, we used the case studies of planar pushing and door opening to integrate visual, haptic, and proprioceptive data.
We show that fusing these sensory inputs in differentiable filters that leverage crossmodal information can provide valuable opportunities for interpretability without sacrificing performance.
This is particularly true for our proposed \texttt{Crossmodal Fusion} architectures, which allow for fine-grained analysis of how each modality contributes to our final state estimate.

%
%
In future work, we will investigate the proposed architectures for a broader set of manipulation tasks and with other state estimation techniques such as Unscented Kalman filters or factor graphs. Furthermore, we want to study how to combine multimodal state estimation with policy learning, which may allow us to select optimal actions for revealing task-relevant information.



\renewcommand*{\bibfont}{\footnotesize}

\printbibliography 

\end{document}